\pdfoutput=1
\documentclass[11pt]{article}

\usepackage[]{acl}

\usepackage{times}
\usepackage{latexsym}
\usepackage{bbold}
\usepackage[T1]{fontenc}

\usepackage[utf8]{inputenc}

\usepackage{microtype}

\usepackage{inconsolata}
\usepackage{booktabs}
\usepackage{enumitem}
\usepackage{titlesec}
\usepackage{multirow}
\usepackage{booktabs} % for nice tables
\usepackage{tablefootnote} % for table footnotes
\usepackage{amsfonts}
\usepackage{float}
\usepackage{amsmath}
\usepackage{graphicx}
\usepackage{subcaption}
\usepackage{colortbl}
\usepackage{bbm}

\usepackage{array}
\newcolumntype{L}[1]{>{\raggedright\let\newline\\\arraybackslash\hspace{0pt}}m{#1}}
\newcolumntype{C}[1]{>{\centering\let\newline\\\arraybackslash\hspace{0pt}}m{#1}}
\newcolumntype{R}[1]{>{\raggedleft\let\newline\\\arraybackslash\hspace{0pt}}m{#1}}

\newcolumntype{C}[1]{>{\centering\arraybackslash}p{#1}}
\definecolor{Gray}{gray}{0.925}
\definecolor{LightCyan}{rgb}{0.88,1,1}

\newcolumntype{a}{>{\columncolor{Gray}}c}
\title{Similarity-Based Domain Adaptation with LLMs}
\author{
    Jie He$^1$ \quad 
    Wendy Zhou$^1$ \quad 
    Jiaoyan Chen$^2$ \quad 
    \textbf{Jeff Z. Pan}$^1$ \; \\
     $^1$ School of Informatics, University of Edinburgh, UK  \\
    $^2$ University of Pittsburgh, Pittsburgh, PA, USA \\
    \normalsize{\texttt{ j.he@ed.ac.uk, s2236454@sms.ed.ac.uk, }} \\ \normalsize{\texttt{xianglli@pitt.edu, j.z.pan@ed.ac.uk}} \\
}

\begin{document}
\maketitle
\begin{abstract}
% Cross-domain sentiment classification aims to address the issue of how to utilize rich labelled data from different but related source domains and unlabeled data from the target domain when the model lacks data from the target domain. Previous research efforts have mainly focused on extracting domain-invariant features and domain-specific perceptual features. However, these approaches all require source domain data to train the source model, which is not only time-consuming but also results in source models that cannot be reused for multiple purposes. In this paper, we introduce the SimCross framework, which leverages the generalization capabilities of Large Language Models (LLMs) to annotate target domain data using source domain data. We then further enhance the performance of the small model in the target domain by fine-tuning it with high-quality target domain labelled data and distilling batch-wise similarity consistency from a large language model. Extensive experiments in cross-domain text classification demonstrate that SimCross achieves superior performance, confirming the effectiveness of our approach.
Cross-domain sentiment classification leverages abundant labeled data from source domains to generalize onto unlabeled target domains. Prior research has primarily focused on learning domain-invariant features across the source and target domains by training a model using both data. In this paper, we evaluate large language models on this task with in-context learning and $k$NN-based prediction using the labeled source data. $k$NN-based models consistently outperform in-context learning, but the performance from smaller models is subpar. We additionally proposed a novel similarity-based knowledge distillation framework. Our extensive experiments on cross-domain sentiment classification reveal that our framework achieves impressive performance, i.e., 2.44\% accuracy improvement compared to the SOTA method.

\end{abstract}
\section{Introduction}
Unsupervised Domain Adaptation (UDA) aims to transfer knowledge from a labeled source domain to an unlabeled target domain. Typically, these domains share common characteristics, such as the positive and negative words in the reviews of two domains (book and movie reviews). Most domain adaptation methods train models to learn domain-invariant features across source and target domains by leveraging only the labeled source domain data~\cite{pmlr-v70-saito17a,  du-etal-2020-adversarial, feng2023twostage}. However, these methods often require model training with labelled source domain data, which is time-consuming and can limit model usage for applications with different source data. The recent generalization ability of large language models (LLMs)~\cite{NEURIPS2020_1457c0d6, NEURIPS2022_b1efde53,workshop2023bloom,openai2023gpt4,touvron2023llama,he2025evaluatingimprovinggraphtext}, raises the question of whether LLMs alone can be used for target domain prediction without retraining.

In this paper, we investigate the effectiveness of large language models in addressing the domain adaptation problem. We experimented with zero-shot and few-shot prompting strategies, i.e in context learning. Furthermore, we experimented with $k$-nearest neighbor ($k$NN) based LLMs by leveraging labeled source data. We showed that the performance of $k$NN-based LLMs improved significantly compared to typical prompting and outperformed multiple previous baselines.

However, smaller language models (SLMs) are still struggling, even with $k$NN-based techniques.. In practice, deploying large models can be expensive and susceptible to efficiency considerations~\cite{openai2023gpt4}. Local models with fewer parameters, namely, SLMs with high performance, are essential in many resource-constrained scenarios. To that end, we proposed a novel similarity-based distillation loss. Our method includes two components: 1) probability-based distillation loss aiming to align LLM-generated probabilities and 2) similarity-based loss between teacher (LLM) and student (SLM) sentence representations.

In summary, we showed the effectiveness of $k$NN-based LLMs in performing the task of cross-domain adaptation. To combat the model efficiency problem, we also present a simple and lightweight framework for distilling cross-domain knowledge to SLMs. Using LLMs, we annotate target domain data via $k$NN augmented annotation and train an SLM using a novel knowledge distillation loss, which not only distills the label distribution of large language models but also provides the SLMs with representation supervision via similarity loss. We show strong knowledge distillation performance improvement using $\text{DistilBERT}_{\text{base}}$ and $\text{BERT}_{\text{base}}$ on eight cross-domain sentiment classification tasks when compared to multiple baselines. We additionally conducted ablation studies of the proposed framework, revealing the importance of similarity-based annotation and distillation on model performance.

\begin{figure*}[t]
    % \begin{center}
    \centering
\includegraphics[%width=0.6\textwidth
        width=14cm, height=5cm
        ]{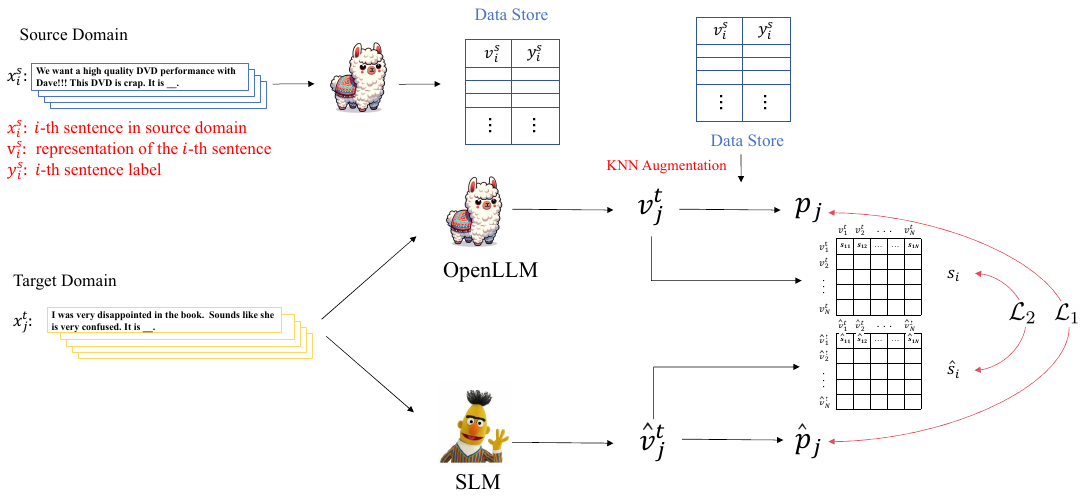}
    % \end{center}
    \caption{Proposed framework Overview. LLM is used to construct the source data store, as shown in the upper left corner. Given target data, we first obtain LLM annotations via $k$NN augmented comparison to the data store, then fine-tune an SLM with two training losses: classification consistency loss $\mathcal{L}_1$ and similarity consistency loss $\mathcal{L}_2$.
}
    \label{main}
% \vspace{-0.5cm}
\end{figure*}
 \section{Related Work}

% \paragraph{Unsupervised Domain Adaptation} 
UDA is commonly used for domain shift problems \cite{ramponi-plank-2020-neural,wright-augenstein-2020-transformer,sheng2021evolving,long-webber-2022-facilitating}. Modern UDA strategies primarily focus on two paradigms. The first is centered around learning domain-invariant features, incorporating methods like pivot-based approaches, domain adversarial training, and class-aware feature self-distillation \cite{sun2019unsupervised, 10.1162/tacl_a_00328, ye-etal-2020-feature, wu-shi-2022-adversarial}. The second paradigm utilizes pretrained models to extract task-agnostic features from the target domain and adapts to its subtleties through language modeling~\cite{Du2021CrossDomainGD,karouzos-etal-2021-udalm,feng2023twostage}. Our work directly evaluates the models' performance without learning or extracting features.
% \paragraph{Knowledge Distillation}

Knowledge Distillation~\cite{hinton2015distilling} transfer knowledge from a high-capacity teacher model to a lower-capacity student model. Knowledge Distillation has been employed for domain adaptation, as discussed by \citet{farahani2020brief}. Previous works have focused on distilling domain-invariant knowledge from the teacher model to the student model \cite{hu-etal-2019-domain,ryu2020knowledge, zhang-etal-2021-matching,long-etal-2020-shallow,long-etal-2024-multi}, which all require extensive training in the source domain.
% Due to the inherent differences between the target and source domains, this often leads to challenges in domain transfer. 
However, our approach of directly utilizing target domain data effectively circumvents this issue, which is similar to symbolic knowledge distillation~\cite{west-etal-2022-symbolic}. We extend the symbolic knowledge distillation with a newly designed similarity-based loss to distil LLM knowledge to smaller models.
% \lorraine{include symbolic knowledge distillation}

% \paragraph{Annotation by Large Language Models}
% Due to the cost and instability of manual labeling, researchers have recently explored using language models for data annotation. \citet{zhang-etal-2021-learning-different}   used the mixup data augmentation framework to assign multiple labels to each example. \citet{wang-etal-2021-want-reduce} investigates using GPT-3 for data annotation. Similarly, \citet{sahu2023promptmix} employs instruction learning to guide large language models to generate and relabel data, while \citet{wang2023lets} extrapolating Errors from Small Models" iteratively guides large language models to generate synthetic datasets with smaller errors compared to real distributions by extrapolating errors from a small model trained on synthesized data against a small real-world validation dataset. Most similar to our work is \cite{saadfalcon2023udapdr}, where large language models are used to generate synthetic queries for reranking tasks. These queries are then used to fine-tune a series of reranking models, which are distilled into a single retriever for adaptation to the target domain. Unlike these methods, we do not require expensive tools like GPT-3 or closed AI; we use open LLMs as effective domain generalization annotation models.

\section{Method}

\begin{table*}[t]

\tiny
\centering
\begin{tabular}{L{5cm}|llllll|C{0.7cm}C{0.7cm}|l}
\toprule
 &
  B $\rightarrow$ C &
  B$\rightarrow$ K &
  C$\rightarrow$ B &
  C$\rightarrow$ K &
  K$\rightarrow$ B &
  K$\rightarrow$ C &
  M$\rightarrow$ S &
  S$\rightarrow$ M &
  Average \\ \midrule
LLM-0                                     & 55.06   & 62.28   & 63.00     & 62.28   & 63.00     & 55.06   & 56.98  & 55.15 & 59.10 \\ 
LLM-ICL                                   & 82.31   & 76.54   & 84.04  & 73.74   & 85.94  & 86.50    & 50.58  & 51.05 & 73.84 \\ 
LLM-$k$NN                                   & \textbf{93.75}   & \textbf{91.51}   & \textbf{90.09}  & \textbf{91.00}      & \textbf{91.26}  & \textbf{91.44}   & \textbf{92.20}   & \textbf{90.00}    & \textbf{91.41} \\ \midrule
\multicolumn{10}{c}{\textbf{$\text{DistilBERT}_{\text{base}}$}}     \\ \midrule
SLM FT-source            & 88.60    & 86.41   & 85.11  & 87.36   & 83.74  & 86.36   & 85.25  & 78.15 & 85.12 \\ 
% SLM train-target           & \textbf{92.15}   & \textbf{90.89}   & \textbf{89.85}  & \textbf{90.89}   & \textbf{89.85}  & \textbf{92.15}   & \underline{88.36}  & 83.1  & \textbf{89.66} \\ \midrule
% SLM train-predicted-target & 90.89   & 87.48   & 86.8   & 88.98   & 86.84  & 89.41   & 86.25  & 81.25 & 87.24 \\ \midrule
SLM-PET~\cite{schick2021exploiting}                   &61.76 & 55.49 & 58.04 & 55.49 & 58.04 & 61.76 & 65.62 & 63.20 & 59.93 \\ 
ICL & 72.23& 74.59 & 71.58 & 75.12 & 72.39 & 72.97 & 58.97 & 58.25 & 69.51 \\
$k$NN  &38.08 &48.60&28.72&30.89&35.41&29.70&39.70&35.55&35.83\\

SLM-KPT~\cite{hu-etal-2022-knowledgeable}                  & 74.13 & 74.81 & 70.11 & 74.97 & 69.60 & 73.92 & 76.84 & 72.70 & 73.39  \\ 
TAMEPT~\cite{feng2023twostage}                 & 90.66 & 87.15 & 86.80 & 88.37 & 85.72 & 87.51 & 88.25 & 83.20 & 87.21  \\ 
Ours                                      
& \textbf{91.31}   & \textbf{89.61}   & \textbf{88.72}  & \textbf{90.35}   & \textbf{88.94}  & \textbf{91.39}   & \textbf{88.85}  & \textbf{83.54} & \textbf{89.09} \\ \midrule
\multicolumn{10}{c}{\textbf{$\text{BERT}_{\text{base}}$}}  \\ \midrule
SLM FT-source           & 88.96   & 87.62   & 86.7   & 85.79   & 86.01  & 87.57   & 87.12  & 82.85 & 86.58 \\ 
% SLM train-target          & 90.74   & \underline{91.58}   & \textbf{91.18}  & \textbf{91.58}   & \textbf{91.18}  & \underline{90.74}   & \underline{89.9}   & \textbf{86.05} & \underline{90.37} \\ \midrule
% SLM train-predicted-target  & 90.18   & 88.53   & 88.46  & 89.53   & 88.14  & 90.54   & 88.03  & 83.65 & 88.38 \\ \midrule
SLM-PET~\cite{schick2021exploiting}                      & 54.21    & 51.75    & 52.46   & 51.75    & 52.46   & 54.21    & 56.12   & 55.20  & 53.52  \\ 
ICL &54.68 & 53.96 & 56.10&53.79& 56.31&54.78&52.11&54.05& 54.47\\
$k$NN  &36.35 &43.81&44.29 & 42.61  &  50.00 &  49.92 & 39.92& 37.05 & 42.99\\

SLM-KPT~\cite{hu-etal-2022-knowledgeable}                     
& 72.53 & 74.10  & 69.52 & 75.89 & 69.35 & 74.10  & 75.85 & 72.65 & 73.00 \\ 
TAMEPT~\cite{feng2023twostage}                                  & 89.56 & 89.46 & 86.79 & 86.82 & 87.75 & 91.03 & 88.68 & 84.45 & 88.07 \\ 
Ours                                      & \textbf{92.49}   & \textbf{91.66 }  & \textbf{90.16}  & \textbf{91.26}   & \textbf{90.05}  & \textbf{92.25}   & \textbf{90.23}  & \textbf{85.95} & \textbf{90.51} \\ \bottomrule
\end{tabular}
\caption{Model performance on different cross-domain situations (source->target) separated by two tasks: Amazon reviews classification and sentiment classification.}
\label{main_table}
\vspace{-0.6cm}
\end{table*}

% Our method revolves around utilizing LLMs for domain generalizable data annotation, which will be used for fine-tuning small models. For data from the target domain, we annotate them by retrieving information from the source domain through LLMs (Section 4.1). After obtaining annotated data for the target domain, we fine-tune the small model and align it with the similarity features of the large model to further improve the performance of the small model (Section 4.2).
% Our method revolves around leveraging the generalization ability of LLM for data annotation in the target domain (\S \ref{sec:method: knn}), and then transferring this domain-generalizable ability to a smaller classification model through distillation. We distill the classification SLM (small language model) in the target domain both with and without the annotated labels as detailed in \S \ref{sec:method:distill}.
% Our method focuses on text classification. 
In this section, we first describe the $k$NN-based LLM data annotation for target domain (\S \ref{sec:method: knn}), and then introduce the novel distillation objectives (loss) 
% when training SLMs 
using annotated data from LLMs 
(\S \ref{sec:method:distill}).

\subsection{$k$NN-based Annotation}
\label{sec:method: knn}
% For cross-domain text classification task, we define a \wendy{labelled} source domain dataset $D_s = \{x_i, y_i\}_{i=1}^m$ with $m$ data samples, and a \wendy{unlabelled} target domain dataset $D_t = \{x_i\}_{i=1}^n$ with $n$ data samples. 
In cross-domain sentiment classification, given labelled source domain $D_s = \{x^s_i, y^s_i\}_{i=1}^m$ with $m$ data samples, the model is tasked to predict labels for the target domain $D_t = \{x^t_i\}_{i=1}^n$ with $n$ examples.
% The goal of cross-domain generalization is to learn a target model $q_t$. 
Considering the domain invariant features across domains,  we 
% use LLMs to compare with source data when annotating target samples. We 
follow the procedure in $k$NN-LM \cite{khandelwal20generalization} to achieve $ k$NN-based prompting. First, 
for each sentence in $D_s$, we append ``\textit{It is [MASK].}'' to the end and obtain the mask token representation from LLM. We build 
a data store for each sentence accordingly as a key-value pair $(v^s_i,y^s_i)$ as shown in Figure~\ref{main}, where the key is the mask token representation from LLM and the value is its original label.
% Specifically, we append ``It is [MASK].`` to the end of each sentence and use the mask token representation as the key $v^s_i$. The value $y^s_i$ is the original class label.

To annotate the $j$th sentence $x^t_j$ in the target domain, We obtain the masked label embedding $v^t_j$ just as we obtain the representation of the source example and use this to query the source data store. We retrieve the $k$ most similar data samples in the source domain by calculating the Euclidean distance $d(v^s_i, v^t_j)$ between the target example and all examples in the data store. Lastly, we annotate $x^t_j$ by calculating the weighted average across $k$ most similar source data sample labels, where the weight is proportional to the similarity (normalized negative distance) between the target sentence and the source sentence as shown in equation (\ref{eq1}):
% \begin{equation}
%     P(y^t|v^t_j) \propto \sum\limits_{(v^s_i,y^s_i)\in K_s}  \mathbb{1}_{y^t=y^s_i} \rm{exp}(-d(v^s_i,v^t_j))
%     \label{eq1}
% \end{equation}
\begin{equation}
    P(y^t|v^t_j) \propto \sum\limits_{(v^s_i,y^s_i=y^t)\in K_s}  \rm{exp}(-d(v^s_i,v^t_j)),
    \label{eq1}
\end{equation}
where $P$ denotes the probability of the target sentence $x^t_j$ having class label $y^t$ and 
% $y^t$ is the possible label space for target domain and
$K_s$ denotes the set of $k$ nearest sentences in the data store of $D_s$. We would annotate the target sentence $x^t_j$ with the probability distribution of all possible labels.
% We end up with a distribution over possible labels for the target sentence where we denote to be $p_j$, note that $y^t_j$ is the label with the maximum probability in $p_j$.
\subsection{Distillation Objectives}
\label{sec:method:distill}
Given the target data annotated by LLM, we propose a novel distillation method to transfer the domain-invariant features and domain-aware features by two losses: classification consistency loss and batch-wise similarity consistency loss.

\noindent\textbf{Classification Consistency Loss} 
The classification consistency loss aims to align the SLM predicted label distribution with the LLM annotated label distribution. For each data sample in $D_t$, where $p_i$ is the LLM annotated label distribution.
% \footnote{We refer $p^t_i$ to $p_i$ in following context and omit the target notation. We will be referring to target domain in the following content. Same applies to $x_i$.}  
% With the KNN-LLM annotated target dataset $D_t = \{x_i, p_i\}_{i=1}^n$, where $p_i$ is the predicted label distribution, we first fine-tune our model by aligning its prediction distribution of the masked label $q_i$ with the LLM annotated distribution $p_i$. 
we calculate the Kullback–Leibler divergence (KL-divergence) between the SLM predicted distribution $\hat{p}_j$ and the LLM annotated distribution $p_j$, and average the KL-divergence over all the data samples in $D_t$.
% For the target model $f_t$, we optimize it by minimizing the following objective:
\
% $\mathcal{L}_1 = -\frac{1}{n}\sum\limits_{i=1}^n p(x_i)log\frac{p(x_i)}{q_t(x_i)}
\begin{equation}
\label{loss1}
    \mathcal{L}_1 = \frac{1}{n}\sum\limits_{i=1}^n D_{KL}(p_i||\hat{p}_i)    
\end{equation}

\noindent\textbf{Similarity consistency loss}
To further transfer the domain-aware features in target domain, we introduce the similarity consistency loss (SimLoss) aligning the sentence representation between LLMs and SLMs. Given a batch of $N$ data samples in $D_t$, 
% ${x_i}, i=1,..,N,$, 
we calculate the cosine similarity between the masked label embedding for each example $x_i^t$ and all the other samples in the same batch. We then normalize all the similarity values to obtain the distribution of batch-wise cosine similarities for $x_i^t$. We conduct the same process for both LLM and SLM and denote the results as $s_i$ and $\hat{s}_i$ as shown in Equation~(\ref{eq:llm_sim}).
% As shown in Equation \ref{eq:llm_sim}, \ref{eq:slm_sim}:
\begin{align}
    s_i &= \left[ s_{1i}, s_{2i}, \dots, s_{Ni} \right] \\
    \hat{s}_i &= \left[ \hat{s}_{1i}, \hat{s}_{2i}, \dots, \hat{s}_{Ni} \right]
 \label{eq:llm_sim}
\end{align}
where 
% $s_i$ represents the batch-wise similarity distribution derived from LLM, $\hat{s}_i$ represents the one from SLM and 
$s_{1i}$ denote the normalized similarity between the first and $i$th example in the batch.
% \begin{equation}
    % S_{t} = [f_{x_i}^Tf_{x_1}, f_{x_i}^Tf_{x_2}, ...,  f_{x_i}^Tf_{x_K}]
    % \label{eq:slm_sim}
% \end{equation}

% Then, we convert the similarity into the form of a probability distribution over neighbor anchors.

The similarity consistency loss is computed as the average of the KL divergence between two probability distributions over all batches.
% \begin{equation}
%     p^i = \frac{\mathrm{exp}(S^i)}{\sum_l^K\mathrm{exp}(S^l)}, \mathrm{for}\ i = 1,2,..., K
% \end{equation}
% \begin{equation}
%     \mathcal{L}_2=D_{KL}(p_g||p_{sm})=\sum\limits_{l=1}^{K}p_g^llog\frac{p_q^l}{p_{sm}^l}
% \end{equation}
\begin{equation}
    \mathcal{L}_2=\frac{1}{N}\sum\limits_{i=1}^{N}D_{KL}(s_i||\hat{s}_i)
\end{equation}

The final distillation loss for training SLM is the sum of classification loss and similarity loss:
$\mathcal{L}=\mathcal{L}_1+\mathcal{L}_2$.
\section{Experiment}
\begin{table*}[h]
\setlength{\belowcaptionskip}{-0.3cm}

\tiny
\centering
\begin{tabular}{L{2cm}llllllC{1cm}C{1cm}l}
\toprule
 &
  B$\rightarrow$ C &
  B$\rightarrow$ K &
  C$\rightarrow$ B &
  C$\rightarrow$ K &
  K$\rightarrow$ B &
  K$\rightarrow$ C &
  M$\rightarrow$ S &
  S$\rightarrow$ M &
  Average \\ \midrule
(1) GT, CE           & 90.74 & 91.58 & 91.18 & 91.58 & 91.18 & 90.74 & 89.90  & 86.05 & 90.37 \\ \midrule
(2) PT, CE           & 90.18 & 88.53 & 88.46 & 89.53 & 88.14 & 90.54 & 88.03 & 83.65 & 88.38 \\ \midrule
(3) GT, CE, SimLoss & 93.27 & 91.96 & 91.76 & 91.96 &91.76 &93.27 &90.28 & 86.38 & 91.31\\ \midrule
(4) PT,CE, SimLoss  & 93.34 & 90.66 & 88.46 & 89.60  & 88.96 & 91.98 & 88.96 & 84.70  & 89.55 \\ \midrule
(5) PT, KL           & 90.94 & 90.50  & 89.06 & 90.38 & 88.55 & 91.85 & 88.77 & 84.65 & 89.31 \\ \midrule
(6) PT, KL, SimLoss & 92.49 & 91.66& 90.16 & 91.26 & 90.05 & 92.25 & 90.23 & 85.95 & 90.51 \\ \bottomrule
\end{tabular}
\caption{Ablation study of SLM BERT performance when training with different labels and losses. GT and PT stand for the ground-truth and the prediction labels. CE and KL stand for Cross-entropy Loss and KL-divergence Loss in Classification Consistency Loss (Eq. \ref{loss1}). SimLoss stands for the similarity consistency Loss.}

\label{tab: ablation}
\end{table*}
% We used Amazon Review Dataset as our main evaluation dataset~\cite{ni_justifying_2019}. 
We used Amazon Reviews Dataset as our main evaluation dataset~\cite{ni_justifying_2019}. 
% since it's a popular English cross-domain sentiment classification benchmark. 
Specifically, we choose the reviews in three domains: \textit{CD (C)}, \textit{Kitchen (K)}, and \textit{Book (B)} and construct 6 cross-domain tasks. In addition, We experimented with two English sentiment analysis datasets, Movie Reviews (M) \cite{pang-lee-2005-seeing} and Stanford Sentiment Treebank V2 (S) \cite{socher-etal-2013-recursive}. We use BERT~\cite{devlin-etal-2019-bert} and DistilBERT~\cite{sanh2020distilbert} as SLMs and Mistral-7B-Instruct-v0.1~\cite{jiang2023mistral} as the LLM~\footnote{More details about the datasets and experimental set up can be found in Appendix~\ref{appendix: exp}.}. 
\subsection{Baselines}
\label{sec: exp: baseline}
% We experiments with several baselines for eight cross-domain sentiment classification setups and demonstrate the effectiveness of our SimCross framework. 
For LLM baselines, we experiment with zero and few-shot (in-context learning) prompting (\textbf{LLM-0 and LLM-ICL})
% . For few-shot prompting
, where we randomly selected 16 examples from the source domain for \textbf{LLM-ICL}. We also report the $k$NN augmented LLM result (as described in \S\ref{sec:method: knn}) (\textbf{LLM-$k$NN}) where $k=16$.  
% \lorraine{what the value of k?} \jieh{k is 16 too}

Additionally, we experiment with several SLM baselines. For unsupervised baselines, we implement PET, a prompt tuning approach where the SLM labels each data sample with a single-word verbalizer (\textbf{SLM-PET})~\cite{schick2021exploiting}. and its extension, KPT where the space of the verbalizer is expanded with external knowledge bases and SLM itself (\textbf{SLM-KPT})~\cite{hu-etal-2022-knowledgeable}. 
% \lorraine{Are these distillation methods? Do they use LLM labels? or any labels?}\jieh{they are not distillation models, we just explored the zero-shot and few-shot prompt learning ability of BERT here} 
For supervised baselines, we train SLM with only source domain labelled data (\textbf{SLM FT-source}), also we reproduce the competitive state-of-the-art two-stage fine-tuning framework (\textbf{TAMEPT}) \cite{feng2023twostage}.

\subsection{Results}

% \lorraine{Lorraine review till here.}
Table \ref{main_table}
shows the performance of our method with baselines on eight cross-domain sentiment classification tasks. 
% that SLM trained with SimCross framework consistently outperforms most LLM-only and all SLM-only baselines across all 8 different cross-domain text classification tasks.\footnote{Details about the experimental set-up and Datasets are in Appendix~\ref{appendix: exp}.} 
% language models.
% cross-domain text classification tasks.
We observe that $k$NN augmented LLM performance is significantly better than the ICL baseline 
% especially for the sentiment analysis task, where LLM-0 and LLM-ICL perform almost randomly. This 
despite using the same number of examples from source domain. For the cross-domain classifications between M and S, LLM-ICL performs almost randomly,  
indicating its instability. 
% and the benefit of selecting the $k$ most similar sentences from the source domain.
Regarding all the SLM-only baselines, our method surpasses the SLM FT-source by an average of 3.9\% with BERT and DistilBERT, highlighting the effectiveness of our proposed distillation approach. Moreover, our accuracy is on average 1.88\% higher with BERT and 2.44\% higher with DistillBERT than the SOTA TAMEPT \cite{feng2023twostage}. 
% without requiring any labelled data in the target domain. 
% compared to the SLM trained on the target, 
% our method even surpasses the SLM trained on target with BERT and 
% acquires a comparable classification accuracy with distilBERT (only 0.57\% behind), while the average accuracy of the target domain training data obtained by $k$NN from the large model is only 91.41\%.
\section{Ablation Studies}
% \subsection{Ablation Studies}
% We conduct three ablation studies.
% concerning the quality of the predicted labels from LLM, the importance of the proposed SimLoss and the advantage of KL-divergence loss over the cross-entropy loss when matching SLM predictions to LLM.
% a detailed analysis of three components: ground truth label vs. predicted label, sim\_loss vs. without\_sim\_loss, and the effectiveness of cross-entropy and KL divergence loss, using the bert model as our pre-trained language model.
First, to evaluate the quality of predicted target labels, we compare the accuracy of SLM trained on ground truth labels with predicted labels. From results (1) vs. (2), (3) vs. (4) in Table \ref{tab: ablation}, we observe that using ground truth labels generally yields better results.
% with the same loss choice. 
However, the difference is not substantial, indicating the high quality of the $k$NN prediction.
% particularly in Amazon cross-entropy setups, when fewer labeled target data are available.
% than predicted labels in experiments with or without sim\_loss.
% This is intuitive, considering the accuracy of predicted labels is only 90.97\%.

Secondly, we found that adding SimLoss consistently improves the performance 
% across various training setups 
through results (1) vs. (3), (2) vs. (4), and (5) vs. (6) in Table \ref{tab: ablation}. The improvement is more pronounced when training on predicted labels, demonstrating the importance of our proposed SimLoss in scenarios with insufficient labelled data.

Finally, from the results (2) vs. (5), (4) vs. (6) in Table \ref{tab: ablation}, we notice that the performance with  KL-divergence Loss is on average better than cross-entropy loss. This could be due to the distribution with KL having better calibration compared to cross entropy which is applied on hard predicted labels~\cite{cuk2023teacher, li-caragea-2023-distilling,he-etal-2022-evaluating,long2024leveraginghierarchicalprototypesverbalizer}. We include an analysis of loss function in Appendix~\ref{appendix: loss}.
% and \citet{li-caragea-2023-distilling}.

% \subsection{Impact of loss type}
% For the calculation of similarity consistency loss, we consider KL loss in the main experimental part, but a native solution is to define a regression loss to close the distance between two similarity distributions:
% \begin{equation*}
% \small
%     \mathcal{L}_2 = \left(\sum\limits_{i=1}^K|S_g^i-S_{sm}^i|^\alpha \right) ^{\frac{1}{\alpha}}, \alpha = 1,2.
% \end{equation*}
% To verify the effectiveness of KL loss, we compare it with this naive option, as shown in Table \ref{tab: loss}. Using KL as a consistency constraint is superior to L1 and L2 loss. L1 loss performs the worst, possibly because the L1 loss uses absolute values as distances, leading to difficulty in optimization.

\section{Case Study}
\begin{table}[t]
\tiny
    \centering
    \begin{tabular}{L{0.95\columnwidth}}
    \toprule
    \textbf{Test example} \\ \midrule
    The liars at CBS don't want you to read this, especially those on 60 minutes.read this book and get well educated. It is \textbf{[great]}.\\ \midrule
    \textbf{Top-3 examples with normal method}\\ \midrule
    1. I have read all quinns books and I think this was the worst written book ever. I...waste of money.
    % read a book a day and... 
    It is \textbf{[bad]}.\\
2. My 6 year old and I enjoyed this book thoroughly. The verse is... we highly recommend this book. It is \textbf{[great]}.\\
3. If I was out of toilet paper, I might buy this book...
% The only use I can think of for a piece of paper with Jan Brewer's 
writing is wiping.
It is \textbf{[bad]}.\\ 
    \midrule 
    \textbf{Top-3 examples with our method}\\ \midrule
    1. Torrey is so profoundly accurate in his writings on...% schizophrenia and manic depression that...
    % that one can safely ignore authors who ridicule or reject his arguments...
    He's that good. 
    It is \textbf{[great]}.\\
2. Mona Charen is the most articulate of her breed! In this book she...% puts the hypocracy of the Left in proper view. Shame is, so many US citizens are in denial!
READ this book and be awakened! It is \textbf{[great]}.\\
3. Full of apparent promise, this confused mess manages to say nothing and go nowhere. A total waste of my time. It is \textbf{[bad]}. \\
\bottomrule
    \end{tabular}
    \caption{Given a sentence from the target test set, the top-3 most similar data samples in source training set.}
    \label{tab:example}
\end{table}

To better understand the improvements brought by SimLoss, we selected a positive class example, correctly predicted with SimLoss but incorrectly without it. We calculate its similarity with the three most semantically similar examples in the training set using sentence embedding. As shown in Table \ref{tab:example}, with SimLoss, two of the top three predicted examples are positive, while without SimLoss training, only the second example is positive. This shows the consistency constraint of SimLoss in the test set, leading to improved model performance.

\section{Conclusion}

In this paper, we propose a novel distillation framework for cross-domain sentiment classification. We annotate target domain data with LLMs without retraining a separate model for each source domain. Having the annotated target data, we distill the domain-invariant and domain-aware features to SLMs like BERT and DistillBERT by our proposed distillation objectives. 
% We transfer the domain generalization knowledge of large language models to . 
Our method defeats the latest SOTA model and achieves comparable results with ground truth fine-tuning, making it a promising method for cross-domain classification with insufficient labelled data.

\section*{Limitation}

Firstly, we only consider one open-source large model, Mistral, in our framework. In the future, we plan to test our method on a broader range of open-source models to understand the domain generalization capabilities of different large language models.

Secondly, we assume that LLMs have rich domain generalization knowledge for specific tasks, but there is no guarantee in real-world applications. This might be addressed through instruction tuning to give LLMs a deeper understanding of the tasks they need to solve.

Thirdly, our research focuses only on sentiment classification tasks. Other natural language processing tasks also face similar challenges, such as cross-domain entity recognition. Therefore, future research should aim to evaluate the applicability of our method to other tasks.

\bibliography{acl_latex}

\begin{thebibliography}{43}
\expandafter\ifx\csname natexlab\endcsname\relax\def\natexlab#1{#1}\fi

\bibitem[{Ben-David et~al.(2020)Ben-David, Rabinovitz, and Reichart}]{10.1162/tacl_a_00328}
Eyal Ben-David, Carmel Rabinovitz, and Roi Reichart. 2020.
\newblock \href {https://doi.org/10.1162/tacl_a_00328} {{PERL: Pivot-based Domain Adaptation for Pre-trained Deep Contextualized Embedding Models}}.
\newblock \emph{Transactions of the Association for Computational Linguistics}, 8:504--521.

\bibitem[{Brown et~al.(2020)Brown, Mann, Ryder, Subbiah, Kaplan, Dhariwal, Neelakantan, Shyam, Sastry, Askell, Agarwal, Herbert-Voss, Krueger, Henighan, Child, Ramesh, Ziegler, Wu, Winter, Hesse, Chen, Sigler, Litwin, Gray, Chess, Clark, Berner, McCandlish, Radford, Sutskever, and Amodei}]{NEURIPS2020_1457c0d6}
Tom Brown, Benjamin Mann, Nick Ryder, Melanie Subbiah, Jared~D Kaplan, Prafulla Dhariwal, Arvind Neelakantan, Pranav Shyam, Girish Sastry, Amanda Askell, Sandhini Agarwal, Ariel Herbert-Voss, Gretchen Krueger, Tom Henighan, Rewon Child, Aditya Ramesh, Daniel Ziegler, Jeffrey Wu, Clemens Winter, Chris Hesse, Mark Chen, Eric Sigler, Mateusz Litwin, Scott Gray, Benjamin Chess, Jack Clark, Christopher Berner, Sam McCandlish, Alec Radford, Ilya Sutskever, and Dario Amodei. 2020.
\newblock \href {https://proceedings.neurips.cc/paper_files/paper/2020/file/1457c0d6bfcb4967418bfb8ac142f64a-Paper.pdf} {Language models are few-shot learners}.
\newblock In \emph{Advances in Neural Information Processing Systems}, volume~33, pages 1877--1901. Curran Associates, Inc.

\bibitem[{Devlin et~al.(2019)Devlin, Chang, Lee, and Toutanova}]{devlin-etal-2019-bert}
Jacob Devlin, Ming-Wei Chang, Kenton Lee, and Kristina Toutanova. 2019.
\newblock \href {https://doi.org/10.18653/v1/N19-1423} {{BERT}: Pre-training of deep bidirectional transformers for language understanding}.
\newblock In \emph{Proceedings of the 2019 Conference of the North {A}merican Chapter of the Association for Computational Linguistics: Human Language Technologies, Volume 1 (Long and Short Papers)}, pages 4171--4186, Minneapolis, Minnesota. Association for Computational Linguistics.

\bibitem[{Du et~al.(2020)Du, Sun, Wang, Qi, and Liao}]{du-etal-2020-adversarial}
Chunning Du, Haifeng Sun, Jingyu Wang, Qi~Qi, and Jianxin Liao. 2020.
\newblock \href {https://doi.org/10.18653/v1/2020.acl-main.370} {Adversarial and domain-aware {BERT} for cross-domain sentiment analysis}.
\newblock In \emph{Proceedings of the 58th Annual Meeting of the Association for Computational Linguistics}, pages 4019--4028, Online. Association for Computational Linguistics.

\bibitem[{Du et~al.(2021)Du, Li, Su, Zhu, and Lu}]{Du2021CrossDomainGD}
Zhekai Du, Jingjing Li, Hongzu Su, Lei Zhu, and Ke~Lu. 2021.
\newblock \href {https://api.semanticscholar.org/CorpusID:235367640} {Cross-domain gradient discrepancy minimization for unsupervised domain adaptation}.
\newblock \emph{2021 IEEE/CVF Conference on Computer Vision and Pattern Recognition (CVPR)}, pages 3936--3945.

\bibitem[{Farahani et~al.(2020)Farahani, Voghoei, Rasheed, and Arabnia}]{farahani2020brief}
Abolfazl Farahani, Sahar Voghoei, Khaled Rasheed, and Hamid~R. Arabnia. 2020.
\newblock \href {http://arxiv.org/abs/2010.03978} {A brief review of domain adaptation}.

\bibitem[{Feng et~al.(2023)Feng, Li, Qin, Xu, and Che}]{feng2023twostage}
Yunlong Feng, Bohan Li, Libo Qin, Xiao Xu, and Wanxiang Che. 2023.
\newblock \href {http://arxiv.org/abs/2304.09820} {A two-stage framework with self-supervised distillation for cross-domain text classification}.

\bibitem[{He et~al.(2022)He, Long, and Xiong}]{he-etal-2022-evaluating}
Jie He, Wanqiu Long, and Deyi Xiong. 2022.
\newblock \href {https://aclanthology.org/2022.codi-1.4/} {Evaluating discourse cohesion in pre-trained language models}.
\newblock In \emph{Proceedings of the 3rd Workshop on Computational Approaches to Discourse}, pages 28--34, Gyeongju, Republic of Korea and Online. International Conference on Computational Linguistics.

\bibitem[{He et~al.(2025)He, Yang, Long, Xiong, Gutierrez-Basulto, and Pan}]{he2025evaluatingimprovinggraphtext}
Jie He, Yijun Yang, Wanqiu Long, Deyi Xiong, Victor Gutierrez-Basulto, and Jeff~Z. Pan. 2025.
\newblock \href {http://arxiv.org/abs/2501.14497} {Evaluating and improving graph to text generation with large language models}.

\bibitem[{Hinton et~al.(2015)Hinton, Vinyals, and Dean}]{hinton2015distilling}
Geoffrey Hinton, Oriol Vinyals, and Jeff Dean. 2015.
\newblock \href {http://arxiv.org/abs/1503.02531} {Distilling the knowledge in a neural network}.

\bibitem[{Hu et~al.(2019)Hu, Wu, Zhao, Guo, Cheng, and Su}]{hu-etal-2019-domain}
Mengting Hu, Yike Wu, Shiwan Zhao, Honglei Guo, Renhong Cheng, and Zhong Su. 2019.
\newblock \href {https://doi.org/10.18653/v1/D19-1558} {Domain-invariant feature distillation for cross-domain sentiment classification}.
\newblock In \emph{Proceedings of the 2019 Conference on Empirical Methods in Natural Language Processing and the 9th International Joint Conference on Natural Language Processing (EMNLP-IJCNLP)}, pages 5559--5568, Hong Kong, China. Association for Computational Linguistics.

\bibitem[{Hu et~al.(2022)Hu, Ding, Wang, Liu, Wang, Li, Wu, and Sun}]{hu-etal-2022-knowledgeable}
Shengding Hu, Ning Ding, Huadong Wang, Zhiyuan Liu, Jingang Wang, Juanzi Li, Wei Wu, and Maosong Sun. 2022.
\newblock \href {https://doi.org/10.18653/v1/2022.acl-long.158} {Knowledgeable prompt-tuning: Incorporating knowledge into prompt verbalizer for text classification}.
\newblock In \emph{Proceedings of the 60th Annual Meeting of the Association for Computational Linguistics (Volume 1: Long Papers)}, pages 2225--2240, Dublin, Ireland. Association for Computational Linguistics.

\bibitem[{Jiang et~al.(2023)Jiang, Sablayrolles, Mensch, Bamford, Chaplot, de~las Casas, Bressand, Lengyel, Lample, Saulnier, Lavaud, Lachaux, Stock, Scao, Lavril, Wang, Lacroix, and Sayed}]{jiang2023mistral}
Albert~Q. Jiang, Alexandre Sablayrolles, Arthur Mensch, Chris Bamford, Devendra~Singh Chaplot, Diego de~las Casas, Florian Bressand, Gianna Lengyel, Guillaume Lample, Lucile Saulnier, Lélio~Renard Lavaud, Marie-Anne Lachaux, Pierre Stock, Teven~Le Scao, Thibaut Lavril, Thomas Wang, Timothée Lacroix, and William~El Sayed. 2023.
\newblock \href {http://arxiv.org/abs/2310.06825} {Mistral 7b}.

\bibitem[{Johnson et~al.(2019)Johnson, Douze, and J{\'e}gou}]{johnson2019billion}
Jeff Johnson, Matthijs Douze, and Herv{\'e} J{\'e}gou. 2019.
\newblock Billion-scale similarity search with {GPUs}.
\newblock \emph{IEEE Transactions on Big Data}, 7(3):535--547.

\bibitem[{Karouzos et~al.(2021)Karouzos, Paraskevopoulos, and Potamianos}]{karouzos-etal-2021-udalm}
Constantinos Karouzos, Georgios Paraskevopoulos, and Alexandros Potamianos. 2021.
\newblock \href {https://doi.org/10.18653/v1/2021.naacl-main.203} {{UDALM}: Unsupervised domain adaptation through language modeling}.
\newblock In \emph{Proceedings of the 2021 Conference of the North American Chapter of the Association for Computational Linguistics: Human Language Technologies}, pages 2579--2590, Online. Association for Computational Linguistics.

\bibitem[{Khandelwal et~al.(2020)Khandelwal, Levy, Jurafsky, Zettlemoyer, and Lewis}]{khandelwal20generalization}
Urvashi Khandelwal, Omer Levy, Dan Jurafsky, Luke Zettlemoyer, and Mike Lewis. 2020.
\newblock {Generalization through Memorization: Nearest Neighbor Language Models}.
\newblock In \emph{International Conference on Learning Representations (ICLR)}.

\bibitem[{Li and Caragea(2023)}]{li-caragea-2023-distilling}
Yingjie Li and Cornelia Caragea. 2023.
\newblock \href {https://doi.org/10.18653/v1/2023.findings-acl.393} {Distilling calibrated knowledge for stance detection}.
\newblock In \emph{Findings of the Association for Computational Linguistics: ACL 2023}, pages 6316--6329, Toronto, Canada. Association for Computational Linguistics.

\bibitem[{Long et~al.(2020)Long, Cai, Reid, Webber, and Xiong}]{long-etal-2020-shallow}
Wanqiu Long, Xinyi Cai, James Reid, Bonnie Webber, and Deyi Xiong. 2020.
\newblock \href {https://aclanthology.org/2020.lrec-1.129/} {Shallow discourse annotation for {C}hinese {TED} talks}.
\newblock In \emph{Proceedings of the Twelfth Language Resources and Evaluation Conference}, pages 1025--1032, Marseille, France. European Language Resources Association.

\bibitem[{Long et~al.(2024)Long, N, and Webber}]{long-etal-2024-multi}
Wanqiu Long, Siddharth N, and Bonnie Webber. 2024.
\newblock \href {https://doi.org/10.18653/v1/2024.findings-acl.500} {Multi-label classification for implicit discourse relation recognition}.
\newblock In \emph{Findings of the Association for Computational Linguistics: ACL 2024}, pages 8437--8451, Bangkok, Thailand. Association for Computational Linguistics.

\bibitem[{Long and Webber(2022)}]{long-webber-2022-facilitating}
Wanqiu Long and Bonnie Webber. 2022.
\newblock \href {https://doi.org/10.18653/v1/2022.emnlp-main.734} {Facilitating contrastive learning of discourse relational senses by exploiting the hierarchy of sense relations}.
\newblock In \emph{Proceedings of the 2022 Conference on Empirical Methods in Natural Language Processing}, pages 10704--10716, Abu Dhabi, United Arab Emirates. Association for Computational Linguistics.

\bibitem[{Long and Webber(2024)}]{long2024leveraginghierarchicalprototypesverbalizer}
Wanqiu Long and Bonnie Webber. 2024.
\newblock \href {http://arxiv.org/abs/2411.14880} {Leveraging hierarchical prototypes as the verbalizer for implicit discourse relation recognition}.

\bibitem[{Loshchilov and Hutter(2019)}]{loshchilov2018decoupled}
Ilya Loshchilov and Frank Hutter. 2019.
\newblock \href {https://openreview.net/forum?id=Bkg6RiCqY7} {Decoupled weight decay regularization}.
\newblock In \emph{International Conference on Learning Representations}.

\bibitem[{Ni et~al.(2019)Ni, Li, and McAuley}]{ni_justifying_2019}
Jianmo Ni, Jiacheng Li, and Julian McAuley. 2019.
\newblock \href {https://doi.org/10.18653/v1/D19-1018} {Justifying {Recommendations} using {Distantly}-{Labeled} {Reviews} and {Fine}-{Grained} {Aspects}}.
\newblock In \emph{Proceedings of the 2019 {Conference} on {Empirical} {Methods} in {Natural} {Language} {Processing} and the 9th {International} {Joint} {Conference} on {Natural} {Language} {Processing} ({EMNLP}-{IJCNLP})}, pages 188--197, Hong Kong, China. Association for Computational Linguistics.

\bibitem[{OpenAI(2023)}]{openai2023gpt4}
OpenAI. 2023.
\newblock \href {http://arxiv.org/abs/2303.08774} {Gpt-4 technical report}.

\bibitem[{Ouyang et~al.(2022)Ouyang, Wu, Jiang, Almeida, Wainwright, Mishkin, Zhang, Agarwal, Slama, Ray, Schulman, Hilton, Kelton, Miller, Simens, Askell, Welinder, Christiano, Leike, and Lowe}]{NEURIPS2022_b1efde53}
Long Ouyang, Jeffrey Wu, Xu~Jiang, Diogo Almeida, Carroll Wainwright, Pamela Mishkin, Chong Zhang, Sandhini Agarwal, Katarina Slama, Alex Ray, John Schulman, Jacob Hilton, Fraser Kelton, Luke Miller, Maddie Simens, Amanda Askell, Peter Welinder, Paul~F Christiano, Jan Leike, and Ryan Lowe. 2022.
\newblock \href {https://proceedings.neurips.cc/paper_files/paper/2022/file/b1efde53be364a73914f58805a001731-Paper-Conference.pdf} {Training language models to follow instructions with human feedback}.
\newblock In \emph{Advances in Neural Information Processing Systems}, volume~35, pages 27730--27744. Curran Associates, Inc.

\bibitem[{Pang and Lee(2005)}]{pang-lee-2005-seeing}
Bo~Pang and Lillian Lee. 2005.
\newblock \href {https://doi.org/10.3115/1219840.1219855} {Seeing stars: Exploiting class relationships for sentiment categorization with respect to rating scales}.
\newblock In \emph{Proceedings of the 43rd Annual Meeting of the Association for Computational Linguistics ({ACL}{'}05)}, pages 115--124, Ann Arbor, Michigan. Association for Computational Linguistics.

\bibitem[{Ramponi and Plank(2020)}]{ramponi-plank-2020-neural}
Alan Ramponi and Barbara Plank. 2020.
\newblock \href {https://doi.org/10.18653/v1/2020.coling-main.603} {Neural unsupervised domain adaptation in {NLP}{---}{A} survey}.
\newblock In \emph{Proceedings of the 28th International Conference on Computational Linguistics}, pages 6838--6855, Barcelona, Spain (Online). International Committee on Computational Linguistics.

\bibitem[{Ryu and Lee(2020)}]{ryu2020knowledge}
Minho Ryu and Kichun Lee. 2020.
\newblock \href {http://arxiv.org/abs/2010.11478} {Knowledge distillation for bert unsupervised domain adaptation}.

\bibitem[{Saito et~al.(2017)Saito, Ushiku, and Harada}]{pmlr-v70-saito17a}
Kuniaki Saito, Yoshitaka Ushiku, and Tatsuya Harada. 2017.
\newblock \href {https://proceedings.mlr.press/v70/saito17a.html} {Asymmetric tri-training for unsupervised domain adaptation}.
\newblock In \emph{Proceedings of the 34th International Conference on Machine Learning}, volume~70 of \emph{Proceedings of Machine Learning Research}, pages 2988--2997. PMLR.

\bibitem[{Sanh et~al.(2020)Sanh, Debut, Chaumond, and Wolf}]{sanh2020distilbert}
Victor Sanh, Lysandre Debut, Julien Chaumond, and Thomas Wolf. 2020.
\newblock \href {http://arxiv.org/abs/1910.01108} {Distilbert, a distilled version of bert: smaller, faster, cheaper and lighter}.

\bibitem[{Schick and Schütze(2021)}]{schick2021exploiting}
Timo Schick and Hinrich Schütze. 2021.
\newblock \href {http://arxiv.org/abs/2001.07676} {Exploiting cloze questions for few shot text classification and natural language inference}.

\bibitem[{Sheng et~al.(2021)Sheng, Li, Zheng, Liang, Dong, Huang, Ji, and Sun}]{sheng2021evolving}
Kekai Sheng, Ke~Li, Xiawu Zheng, Jian Liang, Weiming Dong, Feiyue Huang, Rongrong Ji, and Xing Sun. 2021.
\newblock \href {http://arxiv.org/abs/2103.13561} {On evolving attention towards domain adaptation}.

\bibitem[{Socher et~al.(2013)Socher, Perelygin, Wu, Chuang, Manning, Ng, and Potts}]{socher-etal-2013-recursive}
Richard Socher, Alex Perelygin, Jean Wu, Jason Chuang, Christopher~D. Manning, Andrew Ng, and Christopher Potts. 2013.
\newblock \href {https://aclanthology.org/D13-1170} {Recursive deep models for semantic compositionality over a sentiment treebank}.
\newblock In \emph{Proceedings of the 2013 Conference on Empirical Methods in Natural Language Processing}, pages 1631--1642, Seattle, Washington, USA. Association for Computational Linguistics.

\bibitem[{Sun et~al.(2019)Sun, Tzeng, Darrell, and Efros}]{sun2019unsupervised}
Yu~Sun, Eric Tzeng, Trevor Darrell, and Alexei~A. Efros. 2019.
\newblock \href {http://arxiv.org/abs/1909.11825} {Unsupervised domain adaptation through self-supervision}.

\bibitem[{Touvron et~al.(2023)Touvron, Lavril, Izacard, Martinet, Lachaux, Lacroix, Rozière, Goyal, Hambro, Azhar, Rodriguez, Joulin, Grave, and Lample}]{touvron2023llama}
Hugo Touvron, Thibaut Lavril, Gautier Izacard, Xavier Martinet, Marie-Anne Lachaux, Timothée Lacroix, Baptiste Rozière, Naman Goyal, Eric Hambro, Faisal Azhar, Aurelien Rodriguez, Armand Joulin, Edouard Grave, and Guillaume Lample. 2023.
\newblock \href {http://arxiv.org/abs/2302.13971} {Llama: Open and efficient foundation language models}.

\bibitem[{West et~al.(2022)West, Bhagavatula, Hessel, Hwang, Jiang, Le~Bras, Lu, Welleck, and Choi}]{west-etal-2022-symbolic}
Peter West, Chandra Bhagavatula, Jack Hessel, Jena Hwang, Liwei Jiang, Ronan Le~Bras, Ximing Lu, Sean Welleck, and Yejin Choi. 2022.
\newblock \href {https://doi.org/10.18653/v1/2022.naacl-main.341} {Symbolic knowledge distillation: from general language models to commonsense models}.
\newblock In \emph{Proceedings of the 2022 Conference of the North American Chapter of the Association for Computational Linguistics: Human Language Technologies}, pages 4602--4625, Seattle, United States. Association for Computational Linguistics.

\bibitem[{Wolf et~al.(2020)Wolf, Debut, Sanh, Chaumond, Delangue, Moi, Cistac, Rault, Louf, Funtowicz, Davison, Shleifer, von Platen, Ma, Jernite, Plu, Xu, Le~Scao, Gugger, Drame, Lhoest, and Rush}]{wolf-etal-2020-transformers}
Thomas Wolf, Lysandre Debut, Victor Sanh, Julien Chaumond, Clement Delangue, Anthony Moi, Pierric Cistac, Tim Rault, Remi Louf, Morgan Funtowicz, Joe Davison, Sam Shleifer, Patrick von Platen, Clara Ma, Yacine Jernite, Julien Plu, Canwen Xu, Teven Le~Scao, Sylvain Gugger, Mariama Drame, Quentin Lhoest, and Alexander Rush. 2020.
\newblock \href {https://doi.org/10.18653/v1/2020.emnlp-demos.6} {Transformers: State-of-the-art natural language processing}.
\newblock In \emph{Proceedings of the 2020 Conference on Empirical Methods in Natural Language Processing: System Demonstrations}, pages 38--45, Online. Association for Computational Linguistics.

\bibitem[{Workshop et~al.(2023)Workshop, :, Scao, Fan, and Christopher~Akiki}]{workshop2023bloom}
BigScience Workshop, :, Teven~Le Scao, Angela Fan, and .~Christopher~Akiki, etc. 2023.
\newblock \href {http://arxiv.org/abs/2211.05100} {Bloom: A 176b-parameter open-access multilingual language model}.

\bibitem[{Wright and Augenstein(2020)}]{wright-augenstein-2020-transformer}
Dustin Wright and Isabelle Augenstein. 2020.
\newblock \href {https://doi.org/10.18653/v1/2020.emnlp-main.639} {Transformer based multi-source domain adaptation}.
\newblock In \emph{Proceedings of the 2020 Conference on Empirical Methods in Natural Language Processing (EMNLP)}, pages 7963--7974, Online. Association for Computational Linguistics.

\bibitem[{Wu and Shi(2022)}]{wu-shi-2022-adversarial}
Hui Wu and Xiaodong Shi. 2022.
\newblock \href {https://doi.org/10.18653/v1/2022.acl-long.174} {Adversarial soft prompt tuning for cross-domain sentiment analysis}.
\newblock In \emph{Proceedings of the 60th Annual Meeting of the Association for Computational Linguistics (Volume 1: Long Papers)}, pages 2438--2447, Dublin, Ireland. Association for Computational Linguistics.

\bibitem[{Ye et~al.(2020)Ye, Tan, He, Li, Ng, and Bing}]{ye-etal-2020-feature}
Hai Ye, Qingyu Tan, Ruidan He, Juntao Li, Hwee~Tou Ng, and Lidong Bing. 2020.
\newblock \href {https://doi.org/10.18653/v1/2020.emnlp-main.599} {Feature adaptation of pre-trained language models across languages and domains with robust self-training}.
\newblock In \emph{Proceedings of the 2020 Conference on Empirical Methods in Natural Language Processing (EMNLP)}, pages 7386--7399, Online. Association for Computational Linguistics.

\bibitem[{Zhang et~al.(2021)Zhang, Zhang, Liu, Cheng, and Li}]{zhang-etal-2021-matching}
Bo~Zhang, Xiaoming Zhang, Yun Liu, Lei Cheng, and Zhoujun Li. 2021.
\newblock \href {https://doi.org/10.18653/v1/2021.acl-long.421} {Matching distributions between model and data: Cross-domain knowledge distillation for unsupervised domain adaptation}.
\newblock In \emph{Proceedings of the 59th Annual Meeting of the Association for Computational Linguistics and the 11th International Joint Conference on Natural Language Processing (Volume 1: Long Papers)}, pages 5423--5433, Online. Association for Computational Linguistics.

\bibitem[{Čuk et~al.(2023)Čuk, Senge, Lauri, and Frintrop}]{cuk2023teacher}
Pia Čuk, Robin Senge, Mikko Lauri, and Simone Frintrop. 2023.
\newblock \href {http://arxiv.org/abs/2304.07593} {Teacher network calibration improves cross-quality knowledge distillation}.

\end{thebibliography}

\appendix
\clearpage
\section{Experimental Configuration}
\label{appendix: exp}
\subsection{Models and parameter setting }
\label{appendix: exp: model}
For all of our experiments, we use the off-the-shelf Mistral \cite{jiang2023mistral}, BERT \cite{devlin-etal-2019-bert}, and DistillBERT \cite{sanh2020distilbert} models from Huggingface Transformers \cite{wolf-etal-2020-transformers}. Mistral 7B is used as the large language model\footnote{https://huggingface.co/mistralai/Mistral-7B-v0.1}, while BERT and DistillBERT are the smaller models used. For the $k$NN part, the Datastore is constructed by using Mistral to infer on the training set of the source domain. For nearest neighbor retrieval, we employ FAISS \cite{johnson2019billion}. Similar to \citet{khandelwal20generalization}, we find that L2-based FAISS search yields better results than inner-product searches, so we adopt this setting for our work as well. We set the value of $k$ to 16. For the second phase of training, we use the Adamw optimizer \cite{loshchilov2018decoupled} with learning rates of 5e-5 for Amazon and 5e-6 for SST and MR. The number of epochs is set to 20, with early stopping implemented if there is no validation performance improvement for up to three logging steps. For all training, the batch size is set to 8.

\subsection{Dataset}
\label{appendix:exp:dataset}
We select Amazon Reviews Dataset \cite{ni_justifying_2019} for evaluating our proposed SimCross framework. This dataset is a popular English cross-domain sentiment classification benchmark based on diverse anonymized user reviews across different domains. Specifically, we choose the reviews in three domains: \textit{CD}, \textit{Kitchen}, and \textit{Book} and construct 6 cross-domain tasks. We random sample balanced negative (0) and positive (1) reviews for each domain as shown in Table \ref{tab: dataset}.

Also, we include two more widely used English sentiment analysis datasets, Movie Reviews (MR) and Stanford Sentiment Treebank V2 (SST2).

For each data sample $s$, we fill it into the prompt template: \textit{$s$. It is [MASK]} and we use two label words to verbalize the mask token: terrible/ great.

\begin{table}[h]
\centering
\begin{tabular}{ccccl}
\cline{1-4}
               & Train & Dev  & Test  &  \\ \cline{1-4}
Amazon Book (B)    & 2,000  & 4,000 & 14,000 &  \\ \cline{1-4}
Amazon CD  (C)    & 2,000  & 4,000 & 14,000 &  \\ \cline{1-4}
Amazon Kitchen (K)& 2,000  & 4,000 & 14,000&  \\ \cline{1-4}
Movie Review (M)  & 8,662  & 256  & 2,000  &  \\ \cline{1-4}
SST2 (S)          & 6,920  & 872  & 1,821  &  \\ \cline{1-4}
\end{tabular}
\caption{The number of samples inside Train/ Dev/ Test set for each evaluation dataset. }
\label{tab: dataset}
\end{table}

\section{Analysis: Impact of loss type}
\begin{table*}[h]
\tiny
\centering

\begin{tabular}{L{2cm}ccccccC{1cm}C{1cm}c}
\toprule
 &
  B$\rightarrow$ C &
  B$\rightarrow$ K &
  C$\rightarrow$ B &
  C$\rightarrow$ K &
  K$\rightarrow$ B &
  K$\rightarrow$ C &
  M$\rightarrow$ S&
  S$\rightarrow$ M &
  Average \\ \midrule
$\mathcal{L}^{1}_2$          & 91.56 & 91.54 & 89.63 & 90.58 & 88.66 & 90.64 & 86.77 & 82.40  & 88.97 \\ \midrule
$\mathcal{L}^{2}_2$         & 91.78 & \textbf{92.26} & 89.66 & 90.71 & 89.64 & 91.98 & 86.44 & 83.80  & 89.53 \\ \midrule
$\mathcal{L}^{KL}_2$ & \textbf{92.49} & 91.66 & \textbf{90.16} & \textbf{91.26} & \textbf{90.05} & \textbf{92.25} & \textbf{90.23} & \textbf{85.95} & \textbf{90.51}\\ \bottomrule
\end{tabular}
\caption{Here are the experiment results of different types of SimLoss.  KL stands for KL-divergence loss. }
\label{tab: loss}
\end{table*}

\label{appendix: loss}
For the calculation of similarity consistency loss, we consider KL loss in the main experimental part, but a native solution is to define a regression loss to close the distance between two similarity distributions:
\begin{equation*}
\small
    \mathcal{L}^\alpha_2 = \left(\sum\limits_{i=1}^K|S_g^i-S_{sm}^i|^\alpha \right) ^{\frac{1}{\alpha}}, \alpha = 1,2.
\end{equation*}
To verify the effectiveness of KL loss, we compare it with this naive option, as shown in Table \ref{tab: loss}. Using $\mathcal{L}^{KL}_2$ as a consistency constraint is superior to $\mathcal{L}^{1}_2$  and $\mathcal{L}_{2}^2$  loss.  $\mathcal{L}^{1}_2$  loss performs the worst, possibly because the  $\mathcal{L}^{1}_2$ loss uses absolute values as distances, leading to difficulty in optimization.

\end{document}